\documentclass[letterpaper, 10 pt, conference]{ieeeconf}  

\IEEEoverridecommandlockouts                              

\usepackage{amssymb}
\usepackage{amsmath}
\usepackage{graphicx}
\usepackage{booktabs}
\usepackage{pifont}
\usepackage{array}
\usepackage[dvipsnames]{xcolor}
\usepackage{makecell}
\usepackage{changepage}
\usepackage{hyperref}
\usepackage{mdframed}
\usepackage{multicol} 
\usepackage{subcaption}

\usepackage{epstopdf}

\usepackage{parskip}
\usepackage{lipsum}

\newcommand{\correctA}{\textcolor{JungleGreen}{\ding{108}}}
\newcommand{\correctB}{\textcolor{YellowOrange}{\ding{108}}}
\newcommand{\correctC}{\textcolor{Orchid}{\ding{108}}}
\newcommand{\incorrect}{\textcolor{Gray}{\ding{55}}}
\newcommand{\correctFeedback}{\textcolor{Black}{\ding{108}}}

\newcommand{\vct}{\mathbf}
\newcommand{\vect}[1]{\boldsymbol{#1}}

\usepackage{courier}
\usepackage{tikz}
\usetikzlibrary{calc}

\def\particulartemplate#1{
	\begin{tikzpicture}[overlay, remember picture]
	\draw let \p1 = (current page.west), \p2 = (current page.east) in
	node[minimum width=\x2-\x1, minimum height=0.1cm, rectangle, fill=yellow!35!white, anchor=north west, align=center, text width=\x2-\x1] at ($(current page.north west) + (0,-0.3)$) {\large \textbf{\texttt{#1}} };
	\end{tikzpicture}
}

\usepackage{verbatim}

\IEEEoverridecommandlockouts                              

\title{\LARGE \bf
On the capabilities of LLMs for classifying and segmenting time series of fruit picking motions  into primitive actions
}

\author{Eleni~Konstantinidou$^1$, Nikolaos~Kounalakis$^2$, Nikolaos~Efstathopoulos$^1$ and Dimitrios~Papageorgiou$^{1, 3}$
\thanks{This research is funded by CARPOS project, carried out within the framework of the National Recovery and Resilience Plan Greece 2.0, funded by the European Union - NextGenerationEU (Implementation Body: HFRI). Project Number: 16523.
}
\thanks{$^1$\,E.\;Konstantinidou, N.\;Efstathopoulos and D.\;Papageorgiou are with the Dept.\;of Electrical and Computer Engineering, Hellenic Mediterranean University, 714\,10, Heraklion, Crete,  Greece. {\tt\small dimpapag@hmu.gr}}
\thanks{$^2$\,N.\;Kounalakis is with the Dept.\;of Mechanical Engineering, Hellenic Mediterranean University, 714\,10, Heraklion, Crete,  Greece.}
\thanks{$^3$\,D.\;Papageorgiou is also with the Institute of Computer Science, Foundation for Research and Technology - Hellas, Greece. }}

\begin{document}

\maketitle
\particulartemplate{
	This  paper  is  a  Late Breaking Results report and it will be presented through a poster at the 34th IEEE International Conference on Robot and Human Interactive Communication (ROMAN), 2025 at Eindhoven, the Netherlands.}
\thispagestyle{empty}
\pagestyle{empty}


\begin{abstract} 
\label{sec:abstract}
 Despite their recent introduction to human society, Large Language Models (LLMs) have significantly affected the way we tackle mental challenges in our everyday lives. From optimizing our linguistic communication to assisting us in making important decisions, LLMs, such as ChatGPT, are notably reducing our cognitive load by gradually taking on an increasing share of our mental activities. In the context of Learning by Demonstration (LbD), classifying and segmenting complex motions into primitive actions, such as pushing, pulling, twisting etc, is considered to be a key-step towards encoding a task. In this work, we investigate the capabilities of LLMs to undertake this task, considering a finite set of predefined primitive actions found in fruit picking operations. By utilizing LLMs instead of simple supervised learning or analytic methods, we aim at making the method easily applicable and deployable in a real-life scenario. Three different fine-tuning approaches are investigated, compared on datasets captured kinesthetically, using a UR10e robot, during a fruit-picking scenario.      
\end{abstract}


\section{Introduction} 
\label{sec:Introduction}

    Artificial Intelligence (AI) has become a powerful force that is revolutionizing various fields by developing systems that mimic human intelligence, encompassing this way areas like reasoning, learning and problem-solving \cite{xu2021}. This transformation is primarly due to Large Language Models (LLMs), deep learning models capable of processing and generating natural language data and other content \cite{Suruj}. The evolution of LLMs was driven by the growth of the internet and increasing computing power, which fuelled the vision of Natural Language Processing (NLP) \cite{Wulff2025} through deep learning methods such as Recurrent Neural Networks (RNNs) \cite{Bisong}, and the Transformer model \cite{Vaswani}. This nourished models like BERT \cite{Devlin} and Generative Pre-Trained Transformer (GPT) \cite{Radford}, revolutionizing fields like text generation and language translation. 
        
    From a "robotics"  perspective, segmenting a complex motion into primitive actions is a crucial step in learning human kinematic behaviors through demonstrations. While many works employ Deep Neural Networks (DNN) \cite{Liu2019}\cite{YOU2020105750} for segmentation, and others propose analytic solutions \cite{judd2019a}\cite{Hogan2745} or even AI methods like Temporal Convolutional Networks (TCN) \cite{lea2016temporalconvolutionalnetworksaction} \cite{10.1007/978-3-319-49409-8_7} for explicitly segmenting motion time-series, analytic methods require solid technical skills for action detection and classification, making them hard to deploy in a real-world scenario.

    This work investigates LLM capabilities for classifying primitive actions and segmenting complex fruit-picking motions. In particular, we explore and compare three LLM fine-tuning approaches: a) using only linguistic explanation of the primitive action, b) using a small dataset of representative example motions (e.g., five samples per action) and c) combining both a) and b) approaches. The utilized datasets for both the fine-tuning and  experimental evaluation/comparison of the approaches, consists of kinesthetically captured data, i.e. motion time-series captured by having the human demonstrating the motion directly by moving the robot's end-effector through physical contact.

\section{Motivation and Problem Description}
\label{sec:motivation}

\begin{figure}[ht]
    \centering
    \includegraphics[width=0.5\textwidth]{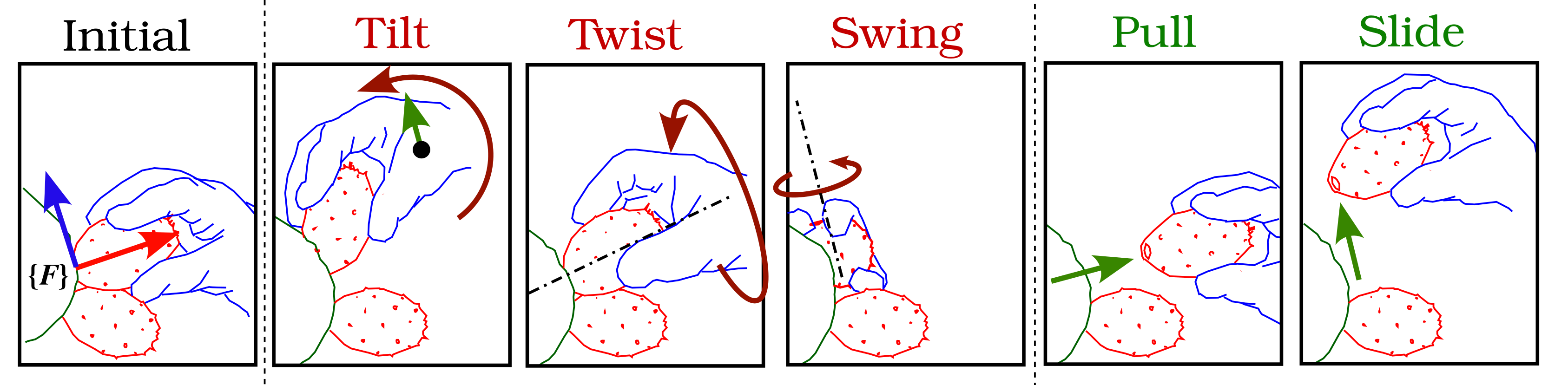}
    \caption{The considered primitive actions}
    \label{fig:primitives}
\end{figure}

The advantages of picking a fruit without utilizing cutting tools span from ensuring the safety of the plant and fruit, to being effective, as in many cases there are no cutting affordances for the insertion of a cutting tool (e.g. in the case of peaches). The complex motion required for detaching a fruit from the branch depends on the type of the plant. However, in most of the cases, this complex motion is composed by a  sequence of simpler primitive actions. As we only account for the detachment of the fruit, these primitive actions are  considered to be the following: \textit{Pull}, \textit{Slide}, \textit{Swing}, \textit{Tilt}, \textit{Twist}, as presented in Fig. \ref{fig:primitives}, with either one of the first two (\textit{Pull} or \textit{Slide}) always appearing  at the end of the complex detachment motion, as they involve a displacement of the fruit from the branch.  

To teach a fruit picking skill to a novice farmer, humans firstly utilize verbal communication, i.e. linguistic explanation. 
An example sentence provided by an experienced farmer to the novice one, to this aim, could be the following: \textit{"To detach the fruit, you firstly have to \underline{twist} it and then \underline{pull} it from the branch."}. In most of the cases, the experienced farmer also provides an example of the fruit picking procedure, by demonstrating it in front of the learner. 
Inspired by the intuition characterizing the human-to-human skill transfer, in this work we aim at investigating the use of Large Language Models (LLMs) for segmenting a complex sequence of primitive actions, in the context of the fruit picking task. Our aim is to exploit these capabilities as a part of Learning by Demonstration (LbD) for a robotic system which will be capable to learn from humans through intuitive means.   

Let us assume the availability of a demonstrated fruit picking motion. Therefore, let $\{F\}$ be the frame at the initial fruit's pose and $\vct{x}(t)\triangleq[\vct{p}^\intercal(t) \; \; \vct{Q}^\intercal(t)]^\intercal\in\mathbb{R}^3\times \mathbb{S}^3$ be the pose of the robot's end-effector with respect to $\{F\}$, where $\vct{p}(t)\in\mathbb{R}^3$ and $\vct{Q}(t)\in\mathbb{S}^3$ are the end-effector's position and orientation in unit quaternion form respectively, with $\mathbb{S}^3$ denoting the unit sphere in $\mathbb{R}^4$. Moreover, let $\vct{v}(t)\triangleq[\dot{\vct{p}}^\intercal(t) \; \; \vect{\omega}^\intercal(t)]^\intercal\in\mathbb{R}^6$ be the generalized velocity of the end-effector with respect to $\{F\}$, with $\dot{\vct{p}}(t), \vect{\omega}(t) \in\mathbb{R}^3$ being its linear and angular velocity respectively. Given that $\vct{x}(t)$ is provided by the forward kinematics of the robot, one can calculate a numerical approximate of $\vct{v}(t)$, based on the numerical differentiation of the pose, as described in the following section. 

Given a time series of the generalized velocity, $\vct{v}(t), t\in[0,T]$, of a complex fruit-picking motion, with respect to the initial pose of the fruit,  with $T\in\mathbb{R}_{>0}$ being the duration of motion, our aim
is to test the effectiveness of an LLM to simultaneously classify and segment the motion into the aforementioned five predefined primitive actions, shown in Fig. \ref{fig:primitives}. 

Notice that the velocity of the end-effector is considered, instead of its pose, in order to facilitate the segmentation by rejecting bias. In other words, by selecting the velocity instead of the pose, each action maintains its characteristics even if it follows after a displacement that could possibly have occurred due to the sequential nature of the complex motion considered.


	

\section{Considered approaches and comparison}
    For classifying and segmenting the time series into primitive actions, a custom Generative Pre-trained Transformer (GPT) was created, based on OpenAI’s GPT-4-turbo model via the ChatGPT Custom GPT framework\footnote{\url{https://help.openai.com/en/articles/8554397-creating-a-gpt}}. 
    A key aspect of the configuration was the emphasis on rule-based classification, where the identification and segmentation of motions were grounded in the predefined kinematic characteristics listed in Table \ref{tab:motion_characteristics}, focusing on the dominant directional patterns observed in the translational and angular velocity components.
    
    Three fine-tuning (learning / explanation) approaches are considered for our case study. The first approach (Approach A) involves only linguistic explanation of the primitive actions. In particular, in this approach the GPT model is only provided with descriptions of the characteristics of the primitive actions with respect to the  expected major axes of motion for each one of them; the characteristics defined in Table \ref{tab:motion_characteristics}. In the second approach (Approach B), a small dataset of representative example motions for each primitive action is provided to the GPT model without however providing any linguistic explanation regarding the characteristics of motion; namely five examples for each primitive action. Finally, Approach C combines both means of Approach A and B, i.e. both linguistic explanation and example datasets are provided to the GPT model.  
\begin{table}[ht]    
    \caption{Kinematic characteristics of primitive actions.} 
    \centering
    \begin{tabular}{ll}
    \toprule
    \textbf{Primitive} & \textbf{Characteristic Behavior} \\
    \midrule
    \textit{Pull}   & Translation along $x$-axis, without significant rotation \\
    \textit{Slide} & Translation on the  $y-z$ plane, without significant rotation \\
    \textit{Swing}  & Translation along the $y$-axis and rotation around the $z$-axis \\
    \textit{Tilt}  & Minimal translation $z$-axis and rotation around $y$-axis \\
    \textit{Twist}  & No translation, but rotation around $x$-axis \\
    \bottomrule \
    \end{tabular}
    \label{tab:motion_characteristics}
    \end{table}
    \vspace{-0.5cm}
    
    In all three approaches, the model was primed to recognize the motion primitives by examining the temporal evolution of the velocity signals and recognizing changes consistent with the prevalent axes defined for each motion type. Segmentation indices were derived by recognizing the earliest significant change in either translational or angular velocity, depending on the defining characteristics of the motion. This ensured that each primitive began at the point where its characteristic movement pattern first emerged. The resulting sequence of motion primitives followed a strict chronological order, determined by their respective start indices.



    In order to allow consistent visualization of velocity profiles and to improve readability of motion transitions, a Python helper function was also provided in the GPT configurations to compute the supremum (maximum absolute value) of the translational and angular velocity components; the script is publicly available\footnote{\href{https://github.com/CSRL-HMU/VerbalDMPcorrections/blob/526440db1fe64ebccb89fe27721f18fc069b9526/motion_classification_segmentation_time_series.ipynb}{GitHub link to script}}. This method was then used to normalize the values of all plots so that an equivalent visual scale is shown regardless of motion magnitude. These resulting plots helped to visually inspect the segmentation boundary validity, making it easier to assess the accuracy and coherence of the identified primitives.

    In line with the recognition and segmentation rules described above, the custom GPT produced an easy readable, structured list of motion primitives with a start index and end index marking a segment border within the time series. An example output format for a complex sequence consisting of three distinct motion primitives, would appear as follows: \emph{twist (Index 0–62)}, \emph{tilt (Index 63–112)}, \emph{pull (Index 113–170)}. This clear and consistent representation supported both interpretability and smooth integration with tasks such as evaluation and visualization.

\subsection{Data Acquisition and Pre-processing}
\begin{figure}[ht]
    \centering
    \includegraphics[width=0.3\textwidth]{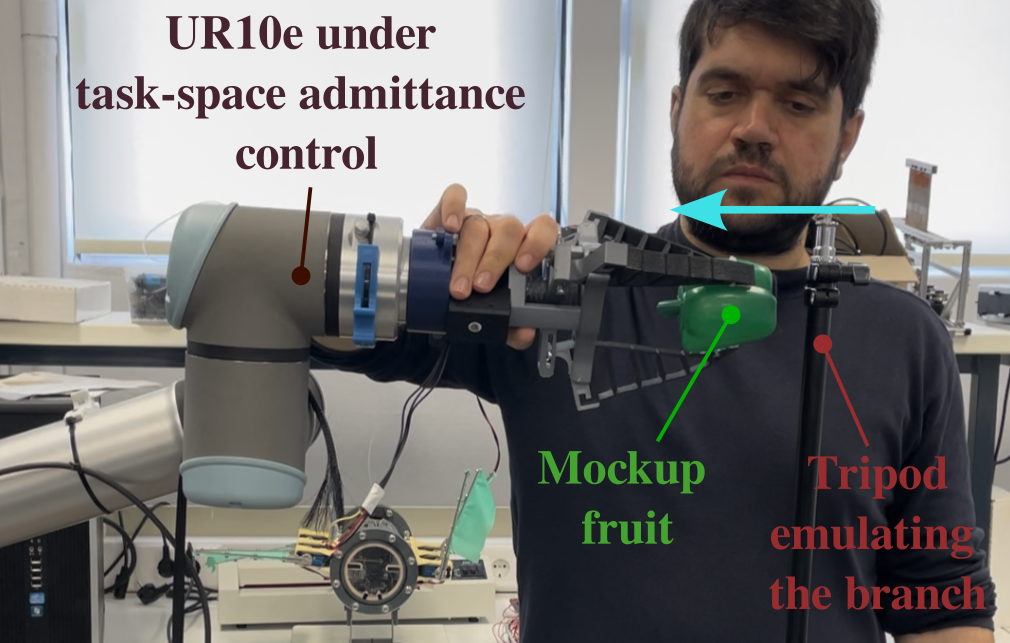}
    \caption{Image from the demonstration of the \textit{pull} primitive.}
    \label{fig:demo_primitives}
\end{figure}

    In order to create a smooth and cooperative human-robot interaction and capture the realistic motion behavior, a task-space admittance control scheme was employed, providing the robot with homogeneous compliance characteristics. For comparison purposes,  a push-button was attached to the handles of the end-effector of the UR10e, serving as a tool for precisely recording the exact moment of motion transitions while capturing the complex motion sequences which was used \underline{only} as ground-truth data for the evaluation. When pressed, it generates a timestamp that effectively marks key transitions. To further emulate a real fruit-picking setup, a mock-up fruit was placed in the robot’s gripper, with the fruit being, at the start of each motion, anchored to a vertical stand that mimicked the presence of a fruit stem. The experimental setup used is depicted in Fig.\ref{fig:demo_primitives}. 

    All experimental data employed in this work were organized into three datasets: training (provided to the model in Approaches B and C as examples), validation (used for assessing the performance of Approach C with feedback)
    and testing, each of which consisted of time-series motion recordings in the same format described above. The training dataset was made up of five demonstrations of each motion primitive separately, i.e. 25 time-series in total.
    As for the testing dataset, its purpose was to assess the model's performance in all the considered approaches. It was made up of complex motion sequences 
    made up of a sequence of the considered  primitive actions. 

    \subsection{Pre-processing}
    The experimental data captured the full $SE(3)$ pose information, i.e. $\vct{x}(t), \forall t\in[0,T]$, provided by the forward kinematics of the robot at $500$Hz. 
    However, one of the important aspects regarding the communication with most of the current GPT models, is the data size provided to the model, with most of the manufacturers imposing data limits or introducing additional costs. Therefore, to resolve the problem of  the high sampling density of the initial time-series (i.e. 500 Hz), the data were interpolated using the Nadaraya-Watson kernel regression method \cite{Nadaraya, watson1964},
     significantly reducing, in this way, the size of the dataset,
     using Gaussian kernels \cite{rasmussen2006}. This non-parametric approach provides an estimate, i.e.  $\hat{y}(t_0)\in\mathbb{R}$, of an arbitrary timeseries $y(t)$, at a given time instance $t_0$,  by computing a kernel-weighted average of observed values $y_i$ corresponding to time instances $t_i$ (i.e. $y_i\triangleq y(t_i)$), as defined by:
    \begin{equation}
    \hat{y}(t) \triangleq \frac{\sum_{i=1}^{n} K(t - t_i)\, y_i}{\sum_{i=1}^{n} K(t - t_i)},
    \end{equation}
    where $K(t) \triangleq\exp\left(-\frac{t^2}{2\sigma^2}\right)\in\mathbb{R}$ is the Gaussian kernel, which assigns weights to each observation at $t_i$ based on its proximity to $t$, with $\sigma^2\in\mathbb{R}_{>0}$ reflecting its variance.
    For the interpolation, 20 kernels per second were considered. Following the interpolation, to thoroughly analyze the dynamic behavior of the system, both translational and angular velocity vectors, i.e. $\dot{\vct{p}}(t)\in\mathbb{R}^3$ and $\vect{\omega}(t)\in\mathbb{R}^3$  respectively,  have been obtained from the non-causal (centered) numerical differentiation of the position and orientation data respectively.  The latter ensures numerical consistency and preserves accuracy across all regions, yielding a discrete approximation of the continuous velocity profile from the estimated derivatives \cite{burden2010numerical}.
    
        
        
    

    To compute the angular velocity vector from the unit quaternion derivative the following formula is used: 
    \begin{equation}
        \vect{\omega}(t) = 2 \mathbf{J}_Q^\intercal(\vct{Q}) \dot{\vct{Q}}(t),
    \end{equation}
where $\vct{J}_{Q}(\vct{Q})\in\mathbb{R}^{4\times 3}$ is the matrix mapping the angular velocities to unit quaternion rates, defined as follows:
\begin{equation}
        \mathbf{J}_Q\vct{(Q)}\triangleq
        \begin{bmatrix}
            -\vect{\epsilon}^\intercal \\[5pt]
            \eta\vct{I}_3 + \vct{S}(\vect{\epsilon})
        \end{bmatrix},
    \end{equation}
with $\vct{S}(.):\mathbb{R}^3\rightarrow\mathbb{R}^{3\times3}$ denoting the skew-symmetric matrix of an $\mathbb{R}^3$ vector and $\eta\in\mathbb{R}, \vect{\epsilon}\in\mathbb{R}^3$ representing the scalar and vector part of the unit quaternion respectively, i.e. $\vct{Q} \triangleq [\eta \; \vect{\epsilon}^\intercal]^\intercal$. More details about the algebra of unit quaternions and its use for describing the orientation of a frame can be found in \cite{Koutras}.

\subsection{Approach A: Explanation through language}
\label{subsec:verbal}
In this approach, the custom GPT model was tasked with identifying and segmenting a series of complex motion sequences using only the kinematic characteristics defined in advance in the model configuration. Importantly, in this approach, the model was not provided with any examples, training data, or data retrieved from motion primitives. Instead, is relied entirely on language-based reasoning, using descriptive rules about each motion behavior (see \autoref{tab:motion_characteristics}).  
The prompt was aimed at looking at velocity profiles for the purpose of classifying motion boundaries, order of occurrence, and segmentation indices. 

\subsection{Approach B: Providing example measurements}
\label{subsec:examples}
Unlike the language-based reasoning of Approach A, this method, introduces an intermediate stage of learning where the GPT is provided with representative motion examples, corresponding to the individual motion primitives: \textit{pull}, \textit{swing}, \textit{slide}, \textit{tilt}, \textit{twist}. In this approach, the specified kinematic characteristics of each motion are not provided to the GPT. Instead, the model has to learn and understand these characteristics implicitly. 

\subsection{Approach C: Combining explanations with examples}
\label{subsec:both}
Approach C integrates the language-based reasoning of Approach A with the example-based reasoning of the Approach B, aiming to leverage the strengths that emerge from the combination of these two approaches. In more details, the model here is provided with both the specific kinematic characteristics and the representative motion examples. 

\begin{figure}[t]
    \centering
     \begin{subfigure}[t]{.25\textwidth}
        \centering
        \includegraphics[scale=0.7]{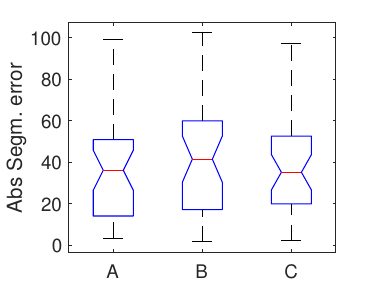}
        \subcaption{Approaches A, B and C}
        \label{fig:ABC_comp}
    \end{subfigure}%
    \begin{subfigure}[t]{.25\textwidth}
         \centering
        \includegraphics[scale=0.7]{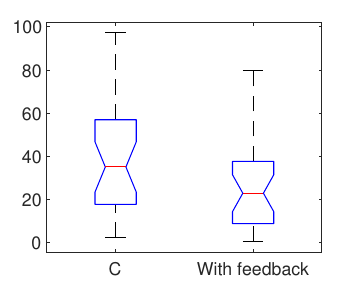}
        \subcaption{With and without feedback}
        \label{fig:setup_with_occ}
    \end{subfigure}
    \caption{Statistical comparison of segmentation error across four approaches.}
    \label{fig:box}
\end{figure}

The performance of the approaches in 20 different complex motion sequences (involving 56 primitive actions) are provided in Table \ref{table:ClassificationResults} and Fig. \ref{fig:box} for the classification and segmentation respectively. Notice that all 20 sequences are considered as the testing dataset for Approaches A, B and C, without the requirement of a ``validation'' dataset.

\begin{table}[ht]  
\caption{Classification of the 20 complex motion sequences, namely  5 validation sequences for the feedback approach and 15 as the testing dataset.}
\label{table:ClassificationResults}
\centering
\begin{adjustwidth}{-0.5cm}{0cm}
\begin{center}
\begin{tabular}{|l|c|c|c|c|}
\cline{1-5}
\makecell{\textbf{Seq.} \\ \textbf{No}} & \makecell{\textbf{1st} \\ \textbf{Segment}} & \makecell{\textbf{2nd} \\ \textbf{Segment}} & \makecell{\textbf{3rd} \\ \textbf{Segment}} & \makecell{\textbf{4th} \\ \textbf{Segment}} \\
\cline{1-5}
1 $\star$ & tilt \correctC\correctFeedback  & 
slide \correctB\correctFeedback
& -- 
& -- \\
\cline{1-5}
2 & tilt \correctC & 
pull \correctFeedback
& -- 
& -- \\
\cline{1-5}
3 & swing \incorrect & 
slide  \correctFeedback
& -- 
& -- \\
\cline{1-5}
4 & swing \incorrect & 
pull  \correctFeedback
& -- 
& -- \\
\cline{1-5}
5  $\star$ & twist \correctA\correctFeedback & 
pull 
\correctFeedback 
& -- 
& -- \\
\cline{1-5}
6 & twist \correctC & 
slide \correctC\correctFeedback
& -- 
& -- \\
\cline{1-5}
7 & twist \correctA\textcolor{YellowOrange}{\ding{108} } & 
tilt \incorrect & 
slide \incorrect 
& --    \\
\cline{1-5}
8 & swing \correctA  & 
tilt \correctFeedback & 
slide \correctFeedback & --  \\
\cline{1-5}
9 & swing \correctC\correctA & 
tilt \correctA & 
pull \correctA\correctFeedback & -- \\
\cline{1-5}
10  $\star$ & twist 
\correctFeedback  & 
tilt 
\correctFeedback & 
pull \correctC\correctFeedback & -- \\
\cline{1-5}
11  & tilt \incorrect & 
twist \correctC &
slide \incorrect & -- \\
\cline{1-5}
12 & tilt \correctFeedback  & 
twist \correctC &
pull \correctA\correctB\correctFeedback & -- \\
\cline{1-5}
13 & swing \correctC &
swing \correctA\textcolor{Orchid}{\ding{108} } & 
slide \incorrect & -- \\
\cline{1-5}
14 $\star$ & swing \correctB\correctC\correctFeedback & 
twist \correctB\correctC\correctFeedback & 
pull \correctC\correctFeedback & -- \\
\cline{1-5}
15 & tilt \correctFeedback & 
swing \correctFeedback & 
slide \incorrect & -- \\
\cline{1-5}
16 $\star$& tilt 
\correctFeedback
& 
swing 
\correctFeedback
& 
pull \correctC\correctFeedback & -- \\
\cline{1-5}
17 & twist \correctFeedback & 
swing \incorrect & 
slide \incorrect & -- \\
\cline{1-5}
18 & twist \incorrect & 
swing \correctB\correctFeedback & 
pull \correctC\correctFeedback & -- \\
\cline{1-5}
19 & twist \correctB\correctFeedback &
tilt \correctA\correctFeedback &
swing \correctA\correctFeedback &
pull \correctA\correctFeedback \\
\cline{1-5}
20 & swing \correctC&
tilt \correctFeedback  &
twist \correctB  & 
slide \incorrect  \\
\cline{1-5}
\end{tabular}
\end{center}
\end{adjustwidth}
\hfill \vspace{0.1cm} \\
\correctA \ Approach A: 11 / 56 (19\%),  \correctB \ Approach B: 8 / 56 (14\%), \\ \correctC \ Approach C: 16 / 56 (28\%), 
\correctFeedback \ Feedback Approach: 19 / 43 (44\%), \\ \incorrect \ None, $\star$ Sequences in which feedback was provided 
\end{table}

\section{Conclusions and the importance of providing feedback to the system}
\label{sec:feedback}
Building upon these findings, the need to provide feedback to the system is evident. The performance variations observed among the three approaches mentioned before indicate the potential for improvement through a targeted feedback process. Towards this direction, five out of the 20 complex motion sequences were used as a ``validation'' dataset for supplying corrective feedback to the system. The results are also shown in Table \ref{table:ClassificationResults} and Fig. \ref{fig:box}. In the context of this experiment, the feedback approach provides some performance improvement by allowing the system to adjust its internal representations and decision processes based on indicated past mistakes and inconsistencies. 

Our next research steps are to increase the performance of the system and apply such methods for providing an intuitive LLM-based human-robot interface for scaling only parts of the learned motion sequence.




\bibliographystyle{IEEEtran}
\bibliography{references} 

\begin{thebibliography}{10}
\providecommand{\url}[1]{#1}
\csname url@rmstyle\endcsname
\providecommand{\newblock}{\relax}
\providecommand{\bibinfo}[2]{#2}
\providecommand\BIBentrySTDinterwordspacing{\spaceskip=0pt\relax}
\providecommand\BIBentryALTinterwordstretchfactor{4}
\providecommand\BIBentryALTinterwordspacing{\spaceskip=\fontdimen2\font plus
\BIBentryALTinterwordstretchfactor\fontdimen3\font minus \fontdimen4\font\relax}
\providecommand\BIBforeignlanguage[2]{{%
\expandafter\ifx\csname l@#1\endcsname\relax
\typeout{** WARNING: IEEEtran.bst: No hyphenation pattern has been}%
\typeout{** loaded for the language `#1'. Using the pattern for}%
\typeout{** the default language instead.}%
\else
\language=\csname l@#1\endcsname
\fi
#2}}

\bibitem{xu2021}
Y.~Xu, X.~Liu, X.~Cao, and et~al., ``Artificial intelligence: A powerful paradigm for scientific research,'' \emph{Innovation (Cambridge)}, vol.~2, no.~4, p. 100179, 2021, © 2021 The Author(s). Published by Elsevier Inc.

\bibitem{Suruj}
A.~Suruj, ``A compact guide to learn large language models,'' 11 2024.

\bibitem{Wulff2025}
P.~Wulff, M.~Kubsch, and C.~Krist, \emph{Natural Language Processing and Large Language Models}.\hskip 1em plus 0.5em minus 0.4em\relax Cham: Springer Nature Switzerland, 2025, pp. 117--142.

\bibitem{Bisong}
E.~Bisong, \emph{Building Machine Learning and Deep Learning Models on Google Cloud Platform: A Comprehensive Guide for Beginners}, 01 2019.

\bibitem{Vaswani}
\BIBentryALTinterwordspacing
A.~Vaswani, N.~Shazeer, N.~Parmar, J.~Uszkoreit, L.~Jones, A.~N. Gomez, L.~Kaiser, and I.~Polosukhin, ``Attention is all you need,'' 2023. [Online]. Available: \url{https://arxiv.org/abs/1706.03762}
\BIBentrySTDinterwordspacing

\bibitem{Devlin}
J.~Devlin, M.-W. Chang, K.~Lee, and K.~Toutanova, ``Bert: Pre-training of deep bidirectional transformers for language understanding,'' 10 2018.

\bibitem{Radford}
A.~Radford, K.~Narasimhan, T.~Salimans, and I.~Sutskever, ``Improving language understanding by generative pre-training,'' 2018.

\bibitem{Liu2019}
X.~Liu, Z.~Deng, and Y.~Yang, ``Recent progress in semantic image segmentation,'' \emph{Artificial Intelligence Review}, vol.~52, no.~2, pp. 1089--1106, 2019.

\bibitem{YOU2020105750}
J.~You, W.~Liu, and J.~Lee, ``A dnn-based semantic segmentation for detecting weed and crop,'' \emph{Computers and Electronics in Agriculture}, vol. 178, p. 105750, 2020.

\bibitem{judd2019a}
K.~Judd, ``Unifying motion segmentation, estimation, and tracking for complex dynamic scenes,'' Ph.D. dissertation, University of Oxford, 2019.

\bibitem{Hogan2745}
N.~Hogan, ``An organizing principle for a class of voluntary movements,'' \emph{Journal of Neuroscience}, vol.~4, no.~11, pp. 2745--2754, 1984.

\bibitem{lea2016temporalconvolutionalnetworksaction}
C.~Lea, M.~D. Flynn, R.~Vidal, A.~Reiter, and G.~D. Hager, ``Temporal convolutional networks for action segmentation and detection,'' 2016.

\bibitem{10.1007/978-3-319-49409-8_7}
C.~Lea, R.~Vidal, A.~Reiter, and G.~D. Hager, ``Temporal convolutional networks: A unified approach to action segmentation,'' in \emph{Computer Vision -- ECCV 2016 Workshops}, G.~Hua and H.~J{\'e}gou, Eds.\hskip 1em plus 0.5em minus 0.4em\relax Cham: Springer International Publishing, 2016, pp. 47--54.

\bibitem{Nadaraya}
E.~A. Nadaraya, ``On estimating regression,'' \emph{Theory of Probability \& Its Applications}, vol.~9, no.~1, pp. 141--142, 1964.

\bibitem{watson1964}
G.~S. Watson, ``Smooth regression analysis,'' \emph{Sankhyā: The Indian Journal of Statistics, Series A (1961–2002)}, vol.~26, no.~4, pp. 359--372, 1964.

\bibitem{rasmussen2006}
C.~E. Rasmussen and C.~K.~I. Williams, \emph{Gaussian Processes for Machine Learning}.\hskip 1em plus 0.5em minus 0.4em\relax Cambridge, Massachusetts: The MIT Press, 2006.

\bibitem{burden2010numerical}
R.~L. Burden and J.~D. Faires, \emph{Numerical Analysis}, 9th~ed.\hskip 1em plus 0.5em minus 0.4em\relax Boston: Brooks/Cole, Cengage Learning, 2010.

\bibitem{Koutras}
L.~Koutras and Z.~Doulgeri, ``Exponential stability of an attitude trajectory tracking controller utilizing unit quaternions,'' 06 2021, pp. 1126--1131.

\end{thebibliography}

\end{document}